%% file: egpaper_for_review.tex
\documentclass[10pt,twocolumn,letterpaper]{article}

\usepackage{cvpr}
\usepackage{times}
\usepackage{epsfig}
\usepackage{graphicx}
\usepackage{amssymb}
\input{math_commands.tex}

\usepackage{caption}
\usepackage{float}
\usepackage{paralist}
\usepackage{makecell}
\usepackage{url}            
\usepackage{booktabs}       
\usepackage{multirow}
\usepackage{threeparttable}
\usepackage{amsfonts}       
\usepackage{nicefrac}       
\usepackage{microtype}      
\usepackage{xcolor}
\usepackage{amsmath, amssymb}
\usepackage{mathtools}
\usepackage{paralist}
\usepackage{wrapfig}
\usepackage[ruled,vlined,noresetcount]{algorithm2e}
\usepackage{url}
\newcommand\disteq{\mathrel{\stackrel{\makebox[0pt]{\mbox{\normalfont\tiny dist.}}}{=}}}
\usepackage[pagebackref=true,breaklinks=true,letterpaper=true,colorlinks,bookmarks=false]{hyperref}

\cvprfinalcopy 

\def\cvprPaperID{5879} 

\ifcvprfinal\fi
\begin{document}

\title{\texttt{Rob-GAN}: Generator, Discriminator, and Adversarial Attacker}

\author{Xuanqing Liu \quad Cho-Jui Hsieh\\
University of California, Los Angeles\\
{\tt\small \{xqliu, chohsieh\}@cs.ucla.edu}
}

\maketitle
\ifcvprfinal
\let\thefootnote\relax\footnotetext{$^*$Project repository: https://github.com/xuanqing94/RobGAN}
\thispagestyle{empty}
\fi

\begin{abstract}

We study two important concepts in adversarial deep learning---adversarial training and generative adversarial network (GAN). 
Adversarial training is the technique used to improve the robustness of discriminator by combining {\bf adversarial attacker} and {\bf discriminator} in the training phase.  
GAN is commonly used for image generation by jointly optimizing {\bf discriminator} and {\bf generator}. 
We show these two concepts are indeed closely related and can be used to strengthen each other---adding a generator to the adversarial training procedure can improve the robustness of  discriminators, and adding an adversarial attack to GAN training can improve the convergence speed and lead to better generators. 
Combining these two insights, we develop a framework called Rob-GAN to jointly optimize generator and discriminator in the presence of adversarial attacks---the generator generates fake images to fool discriminator; the adversarial attacker perturbs real images to fool discriminator, and the discriminator wants to minimize loss under fake and adversarial images. 
Through this end-to-end training procedure, 
we are able to simultaneously improve the convergence speed of GAN training, the quality of synthetic images, and the robustness of discriminator under strong adversarial attacks. 
Experimental results demonstrate that the obtained classifier is more robust than state-of-the-art adversarial training approach~\cite{madry2017towards}, and the generator outperforms SN-GAN on ImageNet-143.

\end{abstract}

\vspace{-10pt}\section{Introduction}
\begin{figure}
    \centering
    \includegraphics[width=0.95\linewidth]{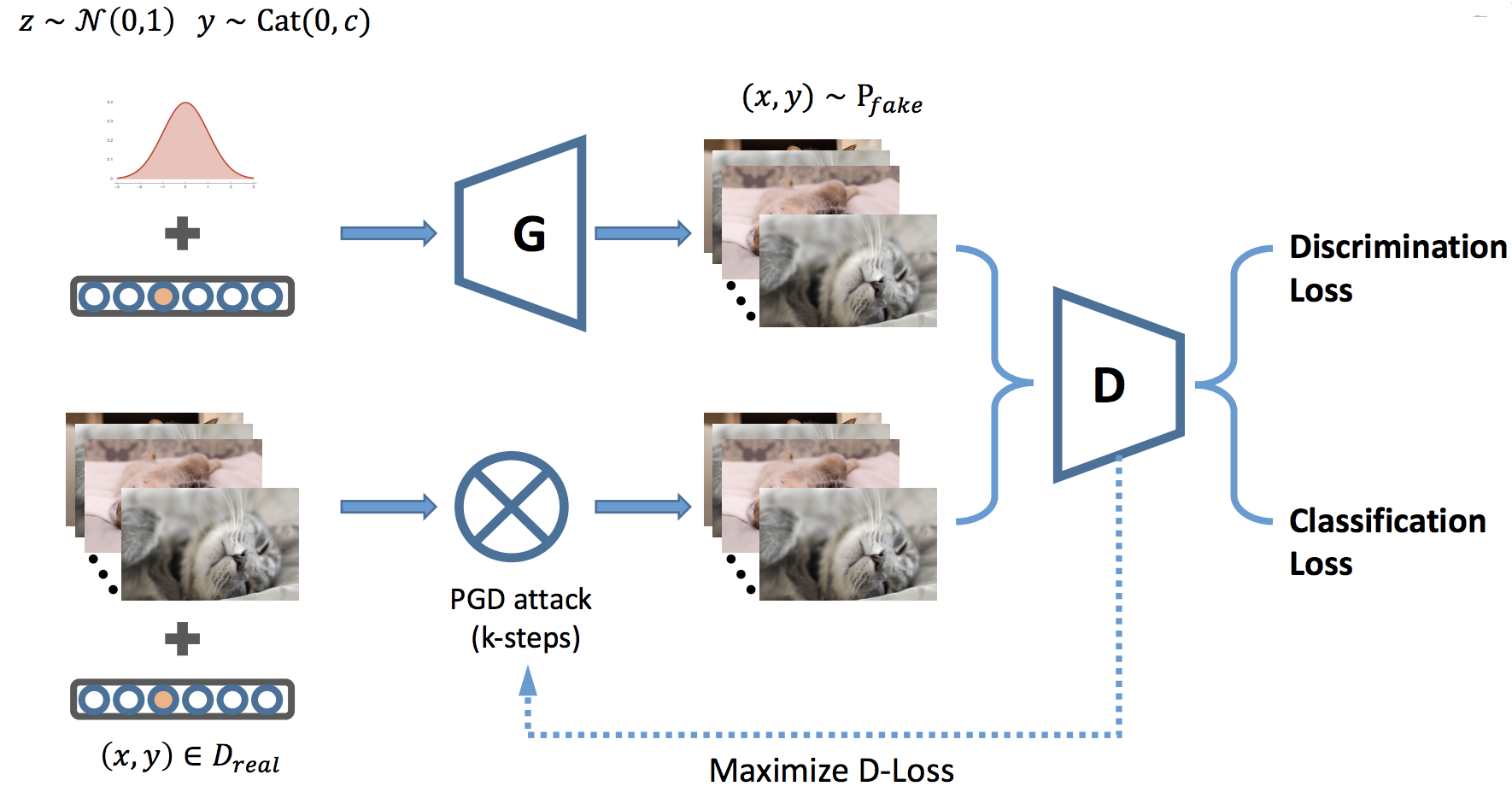}
    \caption{Illustration of the training process. This is similar to the standard GAN training, i.e. alternatively updating the generator $G$ and discriminator $D$ networks. The main difference is that whenever feeding the real images to the $D$ network, we first invoke adversarial attack, so the discriminator is trained with adversarial examples.}
    \label{fig:diagram}
\vspace{-10pt}\end{figure}

Adversarial deep learning has received a significant amount of attention in the last few years. In this paper, we study two important but different concepts---adversarial attack/defense and generative adversarial network (GAN). 
Adversarial attacks are algorithms that find a highly resembled images to cheat a classifier. Training classifiers under adversarial attack (also known as adversarial training) has become one of the most promising ways to improve the robustness of classifiers~\cite{madry2017towards}. On the other hand, GAN is a generative model where the generator learns to convert white noise to images that look authentic to the discriminator~\cite{goodfellow2014generative,odena2017conditional}. 
We show in this paper that they are indeed closely related and can be used to strengthen each other, specifically we have the following key insights: 
\begin{compactenum}
    \item The robustness of adversarial trained classifier can be improved if we have a deeper understanding of the image distribution. Therefore a generator can improve the adversarial training process.
    \item  GAN training can be very slow to reach the equilibrium if the discriminator has a large curvature on the image manifold. Therefore an adversarial trained discriminator can accelerate GAN training. 
\end{compactenum}
Based on these findings, we  managed to accelerate and stabilize the GAN training cycle, by enforcing the discriminator to stay robust on image manifold. At the same time, since data augmentation is used in the robust training process, the generator provides more information about the data distribution. Therefore we get a more robust classifier that generalizes better to unseen data. Our contributions can be summarized as follows:
\begin{compactenum}
    \item We give insights why the current adversarial training algorithm does not generalize well to unseen data. Parallelly, we explain why the GAN training is slow to reach an equilibrium.
    \item We draw a connection between adversarial training and GAN training, showing how they can benefit each other: we can use GAN to improve the generalizability of adversarial training, and use adversarial training to accelerate GAN training and meanwhile make it converge to a better generator. 
    \item We propose a novel framework called \texttt{Rob-GAN}, which integrates generator, discriminator and adversarial attacker as a three-player game. And we also show how to train the framework efficiently in an end to end manner.
    \item We formulate a better training loss for conditional GAN by reformulating the AC-GAN loss. 
    \item We design a series of experiments to confirm all the hypotheses and innovations made in the text. For example, with GAN data augmentation, we can improve the accuracy of state-of-the-art adversarial training method~\cite{madry2017towards} from $29.6\%$ to $36.4\%$ on ResNet18(+CIFAR10) under
    a strong adversarial attack.  Moreover, we observe a $3\sim 7$x speedup in terms of convergence rate, when inserting the adversarial attacker into GAN training cycle. Lastly, our model attains better inception scores on both datasets, compared with the strong baseline (SN-GAN~\cite{miyato2018spectral}).
    
\end{compactenum}

\vspace{-10pt}\paragraph{Notations} Throughout this paper, we denote the (image, label) pair as $(x_i, y_i)$, $i$ is the index of data point; The classifier parameterized by weights $w$ is $f(x; w)$, this function includes the final \texttt{Softmax} layer so the output is probabilities. Loss function is denoted as $\ell(\cdot, \cdot)$. We also define $D(x)$ and $G(z)$ as the discriminator and generator networks respectively. The adversarial example $x_{\text{adv}}$ is crafted by perturbing the original input, i.e. $x_{\text{adv}}=x+\delta$, where $\|\delta\|\le\delta_{\max}$. For convenience, we consider $\ell_{\infty}$-norm in our experiments. The real and fake images are denoted as $x_{\text{real/fake}}$. Note that in this paper ``fake'' images and ``adversarial'' images are different: fake images are generated by generator, while adversarial images are made by perturbing the natural images with small (carefully designed) noise. The training set is denoted as $\mathcal{D}_{\text{tr}}$, with $N_{\mathrm{tr}}$ data points. This is also the empirical distribution. The unknown data distribution is $\mathcal{P}_{\text{data}}$. Given the training set $\mathcal{D}_{\text{tr}}$, we define empirical loss function $\frac{1}{N_{\text{tr}}}\sum_{i=1}^{N_{\text{tr}}}\ell(f(x_i;w),y_i)=\mathop{\mathbb{E}}_{(x,y)\sim\mathcal{D}_{\text{tr}}}\ell(f(x;w),y)$. 

\vspace{-5pt}
\section{Background and Related Work}
\vspace{-5pt}
\subsection{Generative Adaversarial Network}
\label{sec:related_gan}
A GAN has two competing networks with different objectives: in the training phase, the generator $G(z)$ and the discriminator $D(x)$ are evolved in a minimax game, which can be denoted as a unified loss:
\begin{equation}
\label{eq:minimax}
\min_G\max_D\Big\{\mathop{\mathbb{E}}_{x\sim\mathcal{D}_{\text{tr}}}\big[\log D(x)\big]+\mathop{\mathbb{E}}_{z\sim \mathcal{P}_z}\big[\log(1-D(G(z))\big]\Big\},
\end{equation}
where $\mathcal{P}_z$ is the distribution of noise. Unlike traditional machine learning problems where we typically minimize the loss, \eqref{eq:minimax} is harder to optimize and that is the focus of recent literature. Among them, a guideline for the architectures of $G$ and $D$ is summarized in~\cite{radford2015unsupervised}. 
For high resolution and photo-realistic image generation, currently the standard way is to first learn to generate low resolution images as the intermediate products, and then learn to refine them progressively~\cite{denton2015deep,karras2017progressive}. This turns out to be more stable than directly generating high resolution images through a gigantic network. To reach the equilibrium efficiently, alternative loss functions~\cite{arjovsky2017towards,arjovsky2017wasserstein,berthelot2017began,gulrajani2017improved-wgan,unterthiner2017coulomb} are applied and proven to be effective. Among them,  \cite{arjovsky2017towards} theoretically explains why training DCGAN is highly unstable.
Following that work, \cite{arjovsky2017wasserstein} proposes to use Wasserstein-$1$ distance to measure the distance between real and fake data distribution. The resulting network, namely ``Wasserstein-GAN'', largely improves the stability of GAN training. 
Another noteworthy work inspired by WGAN/WGAN-GP is spectral normalization~\cite{miyato2018spectral}. The main idea is to estimate the operator norm $\sigma_{\max}(W)$ of weights $W$ inside layers (convolution, linear, etc.), and then normalize these weights to have $1$-operator norm. 
Because ReLU non-linearity is 1-Lipschitz, if we stack these layers together the whole network will still be 1-Lipschitz, which is exactly the prerequisite to apply Kantorovich-Rubinstein duality to estimate Wasserstein distance.
\par

\subsection{Adversarial attacks and defenses}
\label{sec:intro-attack}

Another key ingredient of our method is adversarial training, originated in~\cite{szegedy2013intriguing} and further studied in~\cite{goodfellow2014explaining}. They found that machine learning models can be easily ``fooled'' by slightly modified images if we design a tiny perturbation according to some ``attack'' algorithms. In this paper we apply a standard algorithm called PGD-attack~\cite{madry2017towards} to generate adversarial examples. Given an example $x$ with ground truth label $y$, PGD computes adversarial perturbation $\delta$ by solving the following optimization with Projected Gradient Descent: 
\begin{equation}
    \label{eq:PGD-attack}
    \delta \coloneqq \argmax_{\|\delta\|\le \delta_{\max}} \ell\big(f(x+\delta; w), y\big),
\end{equation}
where $f(\cdot; w)$ is the network parameterized by weights $w$, $\ell(\cdot, \cdot)$ is the loss function and for convenience we choose $\|\cdot\|$ to be the $\ell_{\infty}$-norm in accordance with~\cite{madry2017towards,athalye2018obfuscated}, but note that other norms are also applicable. Intuitively, the idea of \eqref{eq:PGD-attack} is to find the point $x_{\text{adv}}\coloneqq x+\delta$ within an $\ell_{\infty}$-ball such that the loss value of $x_{\text{adv}}$ is maximized, so that point is most likely to be an adversarial example. In fact, most optimization-based attacking algorithms~(e.g. FGSM~\cite{goodfellow2014explaining}, C\&W~\cite{carlini2017towards}) share the same idea as PGD attack.
\par
Opposite to the adversarial attacks, the adversarial defenses are techniques that make models resistant to adversarial examples. It is worth noting that defense is a much harder task compared with attacks, especially for high dimensional data combined with complex models. Despite that huge amount of defense methods are proposed~\cite{papernot2016distillation,madry2017towards,buckman2018thermometer,ma2018characterizing,guo2018countering,s.2018stochastic,xie2018mitigating,song2018pixeldefend,samangouei2018defensegan}, which can be identified as either random based, projection based, or de-noiser based. In the important overview paper 
~\cite{athalye2018obfuscated,athalye2018robustness}, adversarial training~\cite{madry2017towards} is acknowledged as one of the most powerful defense algorithm, which can be formulated as
\begin{equation}
    \label{eq:Adv-train}
    \min_w\mathop{\mathbb{E}}_{(x,y)\sim\mathcal{P}_{\text{data}}} \Big[\max_{\|\delta\|\le \delta_{\max}} \ell\big(f(x+\delta; w), y\big)\Big],
\end{equation}
where $(x,y)\sim \mathcal{P}_{\text{data}}$ is the (image, label) joint distribution of data, $f(x; w)$ is the network parameterized by $w$, $\ell \big(f(x;w), y\big)$ is the loss function of network (such as the cross-entropy loss). We remark that the ground truth data distribution $\mathcal{P}_{\text{data}}$ is not known in practice, which will be replaced by the empirical distribution.
\par
It is worth noting that one of our contributions is to use GAN to defend the adversarial attacks, which is superficially similar to Defense-GAN~\cite{samangouei2018defense}. However, they are totally different underneath: the idea of Defense-GAN is to project an adversarial example to the space of fake images by minimizing $\ell_2$ distance: $x_{\text{out}}=\mathop{\arg\min}_{G(z)}\|x^{\mathrm{adv}}-G(z)\|^2$, and then making prediction on the output $x_{\text{out}}$. In contrast, our defense mechanism is largely based on adversarial training~\cite{madry2017towards}. Another less related work that applies GAN in adversarial setting is AdvGAN~\cite{xiao2018generating}, where GAN is used to generate adversarial examples. In comparison, we design a collaboration scheme between GAN and adversarial training and both parts are trained jointly.
\section{Proposed Approach}
We propose a framework called \texttt{Rob-GAN} to jointly optimize generator and discriminator in the presence of adversarial attacks---the generator 
generates fake images to fool discriminator; the adversarial attack perturbs real images to fool discriminator, and the discriminator wants to minimize loss under fake and adversarial images (see Fig.~\ref{fig:diagram}). 
In fact, \texttt{Rob-GAN} is closely related to both adversarial training and GAN. If we remove generator, \texttt{Rob-GAN} becomes standard adversarial training method. If we remove adversarial attack, \texttt{Rob-GAN} becomes standard GAN. 
But why do we want to put these three components together? 
Before delving into details, we first present two important motivations: I)
Why can GAN improve the robustness of adversarial trained discriminator? II) Why can adversarial attacker improve the training of GAN? 

We answer I) and II) in Section \ref{sec:gap} and \ref{sec:gan-training}, and then give details of \texttt{Rob-GAN} in Section~\ref{sec:augment}. 

\subsection{Insight I: The generalization gap of adversarial training --- GAN aided adversarial training
}
\label{sec:gap}
\begin{figure*}[htb]
    \centering 
    \vspace{-10pt}
    \scalebox{0.85}{
    \includegraphics[width=0.5\linewidth]{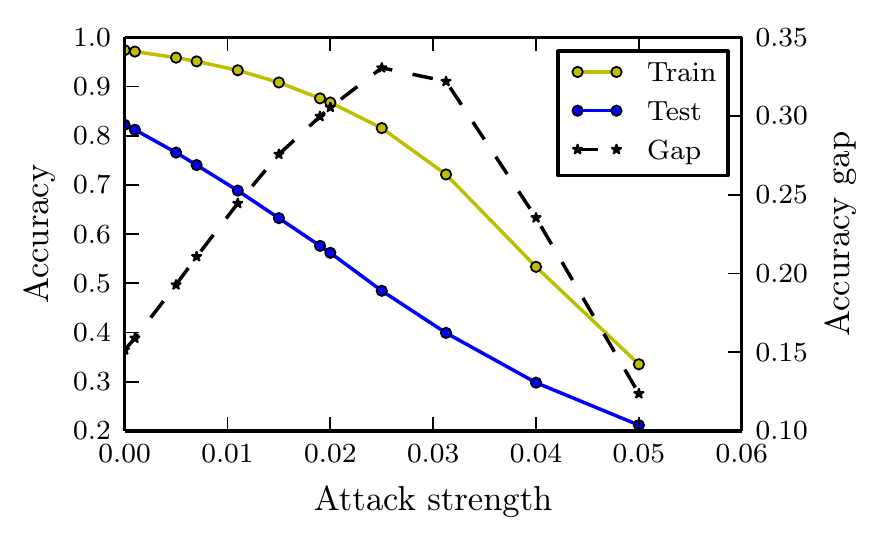}
    \includegraphics[width=0.48\linewidth]{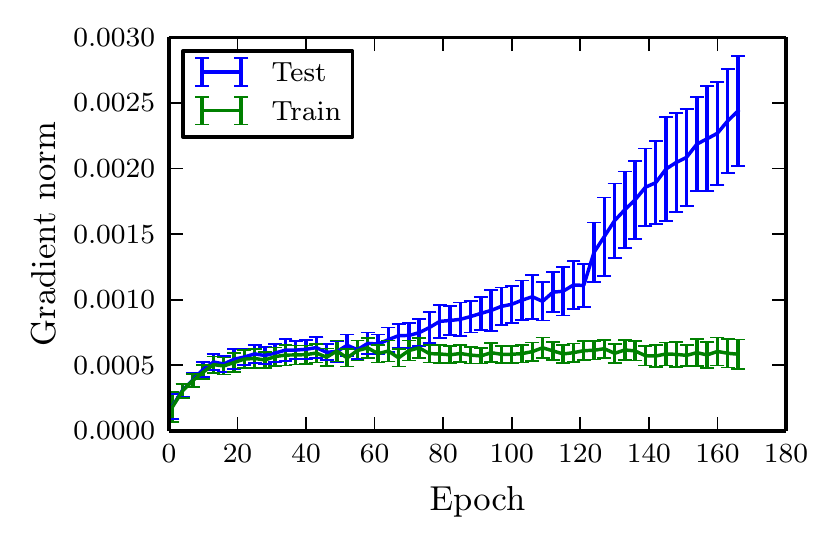}
    }
    \vspace{-7pt}
   \caption{\textit{Left}:  Accuracy under different levels of attack. The model (VGG16) is obtained by adversarial training on CIFAR-10 with the maximum perturbation in adversarial training set as $8/256$. We can observe that: 1) the accuracy gap between training and testing is large, 2) the gap is even larger than the attack strength (after attack strength $\approx0.03$, both training and testing accuracy go down to zero, so the gap also decreases). \textit{Right}: The local Lipschitz value (LLV) measured by gradient norm $\|\frac{\partial}{\partial x_i}\ell\big(f(x_i;w), y_i\big)\|_2$. Data pairs $(x_i, y_i)$ are sampled from the training and testing set respectively. During the training process, LLV on the training set stabilizes at a low level, while LLV on the test set keeps growing.} 
    \label{fig:accuracy-gap}
    \vspace{-5pt}
\end{figure*}
In Sec.~\ref{sec:intro-attack} we listed some works on adversarial defense, and pointed out that adversarial training is one of the most effective defense method to date. However, until now this method has only been tested on small dataset like MNIST and CIFAR10 and it is still an open problem whether adversarial training can scale to large dataset such as ImageNet. Furthermore,
although adversarial training leads to certified robustness on training set (due to the design of the objective function~\eqref{eq:Adv-train}),
the performance usually drops significantly on the test set. 
This means that \textbf{the generalization gap is large} under adversarial attacks (Fig.~\ref{fig:accuracy-gap}~(\textit{Left})).
In other words, despite that it is hard to find an adversarial example near the training data, it is much easier to find one near the testing data. In the following, we investigate the reason behind this huge (and enlarging) generalization gap, and later we will solve this problem with GAN aided adversarial training.
\par
To make our model robust to adversarial distortion, it is desirable to enforce a small local Lipschitz value (LLV) on the underlining data distribution $\mathcal{P}_{\text{data}}$. This idea includes many of the defense methods such as~\cite{cisse2017parseval}. In essence, restricting the LLV can be formulated as a composite loss minimization problem:
\begin{equation}\label{eq:augmented-loss}
    \min_w \mathop{\mathbb{E}}_{(x,y)\sim\mathcal{P}_{\text{data}}}\Big[\ell\big(f(x; w), y\big)+\lambda\cdot\big\| \frac{\partial}{\partial x}\ell\big(f(x;w), y\big)\big\|_2\Big].
\end{equation}
Note that~\eqref{eq:augmented-loss}~can be regarded as the ``linear expansion'' of~\eqref{eq:Adv-train}. In practice we do not know the ground truth data distribution $\mathcal{P}_{\mathrm{data}}$; instead, we use the empirical distribution to replace \eqref{eq:augmented-loss}:
\begin{equation}\label{eq:empirical-augmented-loss}
    \min_w \frac{1}{N_{\mathrm{tr}}}\sum_{i=1}^{N_{\mathrm{tr}}}\Big[\ell\big(f(x_i; w), y_i\big)+\lambda\cdot\big\| \frac{\partial}{\partial x_i}\ell\big(f(x_i;w), y_i\big)\big\|_2\Big],
\end{equation}
where $\{(x_i, y_i)\}_{i=1}^{N_{\mathrm{tr}}}$ are feature-label pairs of the training set. Ideally, if we have enough data and the hypotheses set is moderately large, the objective function in~ \eqref{eq:empirical-augmented-loss} still converges to \eqref{eq:augmented-loss}. However when considering adversarial robustness, we have one more problem to worry about:
\begin{quote}
\textit{Does small LLV in training set automatically generalize to test set?}
\end{quote}
The enlarged accuracy gap shown in Fig.~\ref{fig:accuracy-gap}~(\textit{Left}) implies a negative answer. To verify this phenomenon in an explicit way, we calculate Lipschitz values on samples from training and testing set separately (Fig.~\ref{fig:accuracy-gap}~(\textit{Right})). We can see that similar to the accuracy gap, the LLV gap between training and testing set is equally large. Thus we conclude that \textbf{although adversarial training controls LLV around training set effectively, this property does not generalize to test set}. Notice that our empirical findings do not contradict the certified robustness of adversarial training using generalization theory (e.g.~\cite{sinha2017certifiable}), which can be loose when dealing with deep neural networks. 
\par
The generalization gap can be reduced if we have a direct access to the whole distribution  $\mathcal{P}_{\text{data}}$, instead of approximating it by limited training data. This leads to our first motivation:
\begin{quote}
\textit{Can we use GAN to learn $\mathcal{P}_{\text{data}}$ and then perform the adversarial training process on the learned distribution?} 
\end{quote}
If so, then it becomes straightforward to train an even more robust classifier. Here we give the loss function for doing that, which can be regarded as composite robust optimization on both original training data and GAN synthesized data:
\begin{equation}\label{eq:aug_semi_loss}
\begin{aligned}
    &\min_{w} \mathcal{L}_{\mathrm{real}}(w, \delta_{\max})+\lambda\cdot\mathcal{L}_{\mathrm{fake}}(w, \delta_{\max}),\\ &\mathcal{L}_{\mathrm{real}}(w, \delta_{\max})\triangleq \frac{1}{N_{\mathrm{tr}}}\sum_{i=1}^{N_{\mathrm{tr}}} \max_{\|\delta_i\|\le\delta_{\max}}\ell\big(f(x_i+\delta_i; w); y_i\big),\\
    &\mathcal{L}_{\mathrm{fake}}(w, \delta_{\max})\triangleq \mathop{\mathbb{E}}_{(x, y)\sim\mathcal{P}_{\mathrm{fake}}}\max_{\|\delta\|\le\delta_{\max}}\ell\big(f(x+\delta; w); y\big).
\end{aligned}
\end{equation}
Again the coefficient $\lambda$ is used to balance the two losses. To optimize the objective function \eqref{eq:aug_semi_loss}, we adopt the same stochastic optimization algorithm as adversarial training. That is, at each iteration we draw samples from either training or synthesized data, find the adversarial examples, and then calculate stochastic gradients upon the adversarial examples. We will show the experimental results in Sec. \ref{sec:experiment}. 

\subsection{Insight II: Accelerate GAN training by robust discriminator}
\label{sec:gan-training}
If even a well trained deep classifier can be easily ``cheated'' by adversarial examples, so can the others. Recall in conditional GANs, such as AC-GAN~\cite{odena2017conditional}, the discriminator should not only classify real/fake images but also assign correct labels to input images. Chances are that, if the discriminator is not robust enough to the adversarial attacks, then the generator could make use of its weakness and ``cheat'' the discriminator in a similar way. Furthermore, even though the discriminator can be trained to recognize certain adversarial patterns, the generator will find out other adversarial patterns easily, so the minimax game never stops. Thus we make the following hypothesis:
\begin{quote}
    \textit{Fast GAN training relies on robust discriminator.}
\end{quote}
Before we support this hypothesis with experiments, we briefly review the development of GANs: the first version of GAN objective~\cite{goodfellow2014generative} is unstable to train, WGAN~\cite{arjovsky2017wasserstein,gulrajani2017improved} adds a gradient regularizer to enforce the discriminator to be globally 1-Lipschitz continuous. Later on, SN-GAN~\cite{miyato2018spectral} improves WGAN by replacing gradient regularizer with spectral normalization, again enforcing 1-Lipschitz continuity globally in discriminator. We see both methods \textit{implicitly} make discriminator to be robust against adversarial attacks, because a small Lipschitz value (e.g. 1-Lipschitz) enables stronger invariance to adversarial perturbations.
\par
\begin{figure}[htb]
    \centering
    \includegraphics[width=0.8\linewidth]{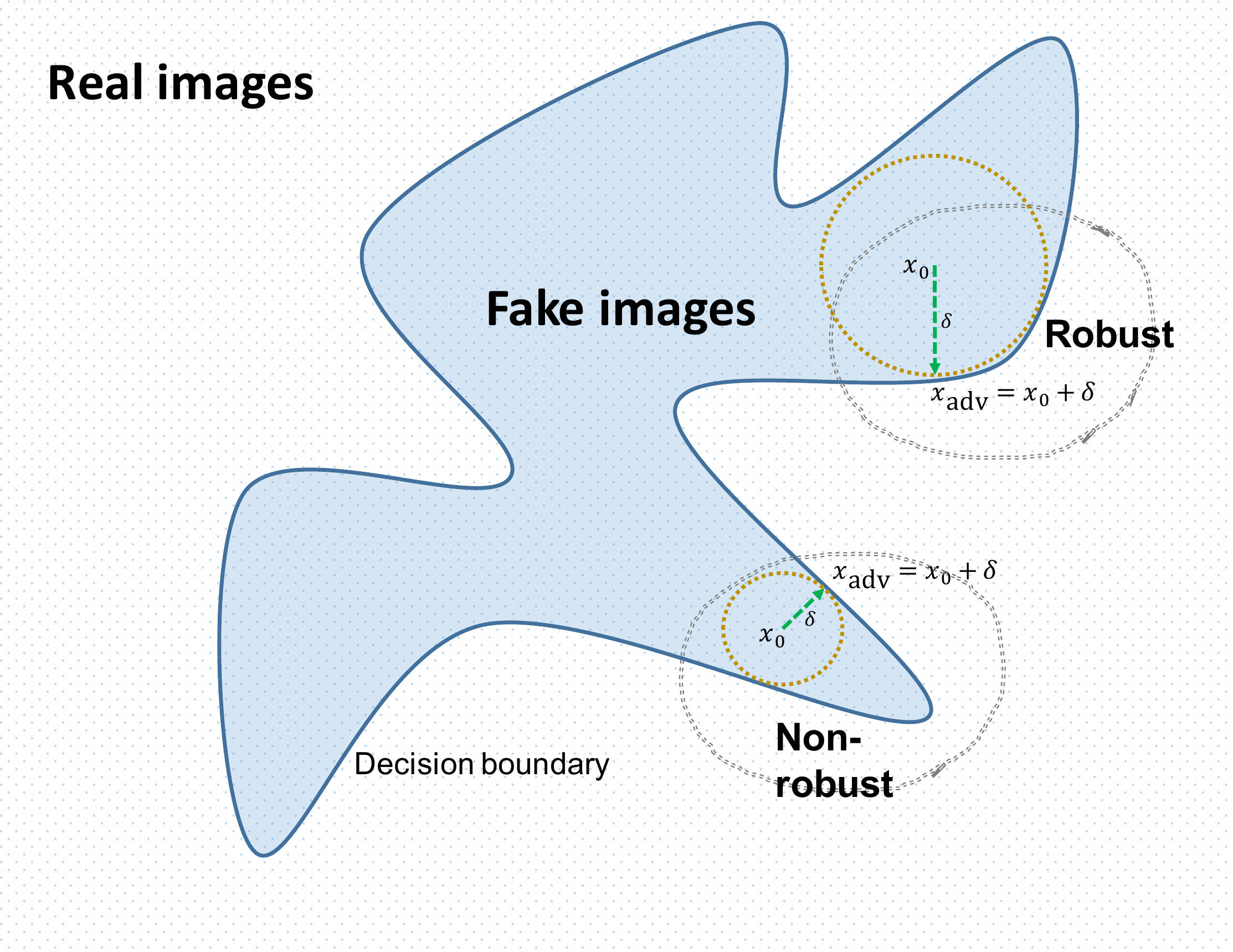}
    \vspace{-8pt}
\caption{Comparing robust and non-robust discriminators, for simplicity, we put them together into one graph. Conceptually, the non-robust discriminator tends to make all images close to the decision boundary, so even a tiny distortion $\delta$ can move a fake image $x_0$ to across the decision boundary and leads to a mis-classification: $x_{\text{adv}}=x_0+\delta$. In contrast, such $\delta$ is expected to be much larger for robust discriminators.}
    \label{fig:boundary}
\end{figure}

Despite the success along this line of research, we wonder if a weaker but smarter regularization to the discriminator is possible. After all, if the regularization effect is too strong, then the model expressiveness will be restricted. Concretely, instead of a strict \textbf{one}-Lipschitz function \textbf{globally}, we require a \textbf{small local } Lipschitz value on image manifold. As we will see, this can be done conveniently through adversarial training to the discriminator. In this way, we can draw a connection between the robustness of discriminator and the learning efficiency of generator, as illustrated in Fig.~\ref{fig:boundary}.
\par
As one can see in Fig.~\ref{fig:boundary}, if a discriminator $D(x)$ has small LLV (equivalently, small $\|D'(x)\|$), then we know $D(x+\delta)\approx D(x)+D'(x)\cdot \delta\approx D(x)$ for a ``reasonably'' small $\delta$. In other words, for a robust discriminator, the perturbed fake image $x_{\text{adv}}=x_0+\delta$ is unlikely to be misclassified as real image, unless $\delta$ is large. Different from  the setting of adversarial attacks~\eqref{eq:PGD-attack}, in GAN training, the ``attacker'' is now a generator network $G(z; w)$ parameterized by $w\in \mathbb{R}^d$. Suppose at time $t$, the discriminator can successfully identify fake images, or equivalently $D(G(z;w^t))\approx 0$ for all $z$, then at time $t+1$ what should the generator do to make $D(G(z;w^{t+1}))\approx 1$? We can develop the following bound by assuming the Lipschitz continuity of $D(x)$ and $G(z; w)$, 
\begin{equation}
    \label{eq:bound-change}
    \begin{aligned}
    1\approx &D(G(z;w^{t+1}))-D(G(z;w^t))\\
    \lessapprox &\|D'\big(G(z;w^t)\big)\|\cdot \|G(z;w^{t+1})-G(z;w^t)\|\\
    \lessapprox 
    &\|D'\big(G(z;w^t)\big)\|\cdot \|\frac{\partial}{\partial w}G(z;w^t)\|\cdot \|w^{t+1}-w^t\|\\
    \le
    &L_DL_G\|w^{t+1}-w^t\|,
    \end{aligned}
\end{equation}
where $L_{D,G}$ indicates the Lipschitz constants of discriminator and generator. As we can see, the update of generator weights is inversely proportional to $L_D$ and $L_G$: $\|w^{t+1}-w^t\|\propto \frac{1}{L_DL_G}$. If the discriminator is lacking robustness, meaning $L_D$ is large, then the generator only needs to make a small movement from the previous weights $w^t$, making the convergence very slow. This validates our hypothesis that \textit{fast GAN training relies on robust discriminator}. In the experiment section, we observe the same phenomenon in all two experiments, providing a solid support for this hypothesis.

\subsection{\texttt{Rob-GAN}: Adversarial training on learned image manifold}
\label{sec:augment}

Motivated by Sec.~\ref{sec:gap} and \ref{sec:gan-training}, we propose a system that combines generator, discriminator, and adversarial attacker into a single framework.
Within this framework, we conduct end-to-end training for both generator and discriminator: the generator feeds fake images to the discriminator; meanwhile real images sampled from training set are preprocessed by PGD attacking algorithm before sending to the discriminator.
 The network structure is illustrated in Fig.~\ref{fig:diagram}.


\begin{figure*}[htb]
    \centering
    \includegraphics[width=0.65\linewidth]{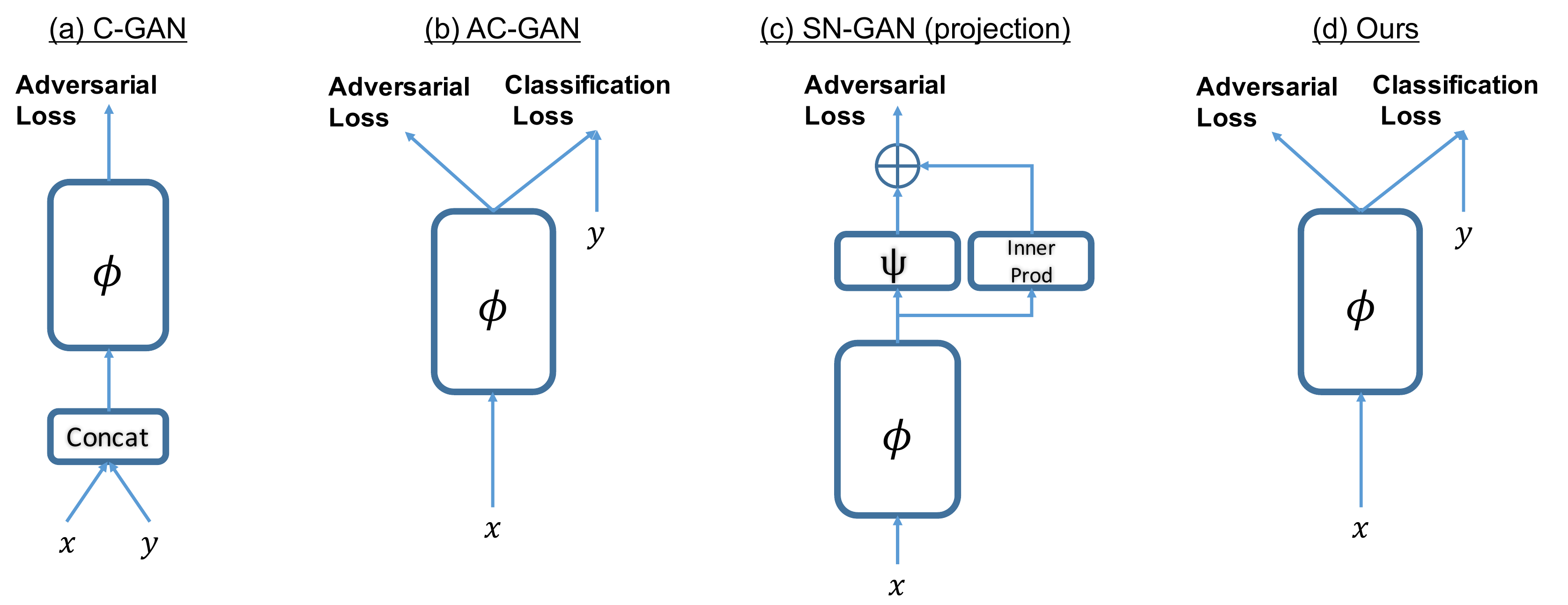}
    \vspace{-5pt}\caption{Comparing the architectures of discriminators. Our architecture is similar to AC-GAN~\cite{odena2017conditional}, but they are different in loss functions, if one compares \eqref{eq:loss_acgan_origin} with \eqref{eq:L_C_splitted}. $(x, y)$ is (image, label) pair, $\phi$ and $\psi$ denote different network blocks. Recall in AC-GAN and our architecture, the discriminator has two branches, one is for discriminating ``real/fake'' images and  the other is for  classification.}
    \vspace{-10pt}
    \label{fig:discriminators}
\end{figure*}
\vspace{-10pt}
\paragraph{Discriminator and the new loss function:} The discriminator could have the standard architecture like AC-GAN. At each iteration, it discriminates real and fake images. When the ground truth labels are available, it also predicts the classes. In this paper, we only consider the conditional GANs proposed in~\cite{mirza2014conditional,odena2017conditional,miyato2018cgans}, and  their architectural differences are illustrated in Fig.~\ref{fig:discriminators}. Among them we simply choose AC-GAN, despite that SN-GAN (a combination of spectral normalization~\cite{miyato2018spectral} and projection discriminator~\cite{miyato2018cgans}) performs much better in their paper. The reason we choose AC-GAN is that SN-GAN's discriminator relies on the ground truth labels and their objective function is not designed to encourage high classification accuracy. But surprisingly, even though AC-GAN is beaten by SN-GAN by a large margin, after inserting the adversarial training module, the performance of AC-GAN matches or even surpasses the SN-GAN, due to the reason discussed in Sec.~\ref{sec:gan-training}. 

We also improved the loss function of AC-GAN. Recall that the original loss in~\cite{odena2017conditional} defined by discrimination likelihood $\mathcal{L}_S$ and classification likelihood $\mathcal{L}_C$:
\begin{equation}
    \label{eq:loss_acgan_origin}
    \begin{aligned}
    \mathcal{L}_S&=\mathbb{E}[\log\mathbb{P}(S=\text{real}|X_{\text{real}})]+\mathbb{E}[\log\mathbb{P}(S=\text{fake}|X_{\text{fake}})]\\
    \mathcal{L}_C&=\mathbb{E}[\log\mathbb{P}(C=c|X_{\text{real}})]+\mathbb{E}[\log\mathbb{P}(C=c|X_{\text{fake}})],
    \end{aligned}
\end{equation}
where $X_{\text{real/fake}}$ are any real/fake images, $S$ is the discriminator output,  and $C$ is the classifier output. Based on~\eqref{eq:loss_acgan_origin}, the goal of discriminator is to maximize $\mathcal{L}_S+\mathcal{L}_C$ while generator aims at maximizing $\mathcal{L}_C-\mathcal{L}_S$. According to this formula, both $G$ and $D$ are trained to increase $\mathcal{L}_C$, which is problematic because even if $G(z; w)$ generates bad images, $D(x)$ has to struggle to classify them (with high loss), and in such case the corresponding gradient term $\nabla \mathcal{L}_C$ can contribute uninformative direction to the discriminator. To resolve this issue, we split $\mathcal{L}_C$ to separate the contributions of real and fake images,
\begin{equation}
    \label{eq:L_C_splitted}
    \begin{aligned}
    \mathcal{L}_{C_1}=\mathbb{E}[\log\mathbb{P}(C=c|X_{\text{real}})]\\ \mathcal{L}_{C_2}=\mathbb{E}[\log\mathbb{P}(C=c|X_{\text{fake}})],
    \end{aligned}
\end{equation}
then discriminator maximizes $\mathcal{L}_S+\mathcal{L}_{C_1}$ and generator maximizes $\mathcal{L}_{C_2}-\mathcal{L}_S$. The new objective function ensures that discriminator only focuses on classifying real images and discriminating real/fake images, and the classifier branch will not be distracted by the fake images.



\paragraph{Generator:} Similar to the traditional GAN training, the generator is updated on a regular basis to mimic the distribution of real data. This is the key ingredient to improve the robustness of classification task: as shown in Sec.~\ref{sec:gap}, model from adversarial training performs well on training set but is vulnerable on test set. Intuitively, this is because during adversarial training, the network only ``sees'' adversarial examples residing in the small region of all training samples, whereas the rest images in the data manifold are undefended. Data augmentation is a natural way to
resolve this issue, but traditional data augmentation methods like image jittering, random resizing, rotation, etc.~\cite{krizhevsky2012imagenet,halevy2009unreasonable,tokozume2017learning,zhang2017mixup,inoue2018data} are all simple geometric transforms, they are useful but not effective enough: even after random transforms, the total number of training data is still much fewer than required. Instead, our system has unlimited samples from generator to provide a continuously supported probability density function for the adversarial training. Unlike traditional augmentation methods, if the equilibrium in \eqref{eq:minimax} is reached, then we can show that the solution of \eqref{eq:minimax} would be~$\mathcal{P}_{\text{fake}}(z)\disteq\mathcal{P}_{\text{real}}$~\cite{goodfellow2014generative}, and therefore the classifier can be trained on the ground truth distribution~$\mathcal{P}_{\mathrm{real}}$.

\paragraph{Fine-tuning the classifier:}
\begin{figure}[htb]
    \centering
    \vspace{-15pt}
    \includegraphics[width=0.7\linewidth]{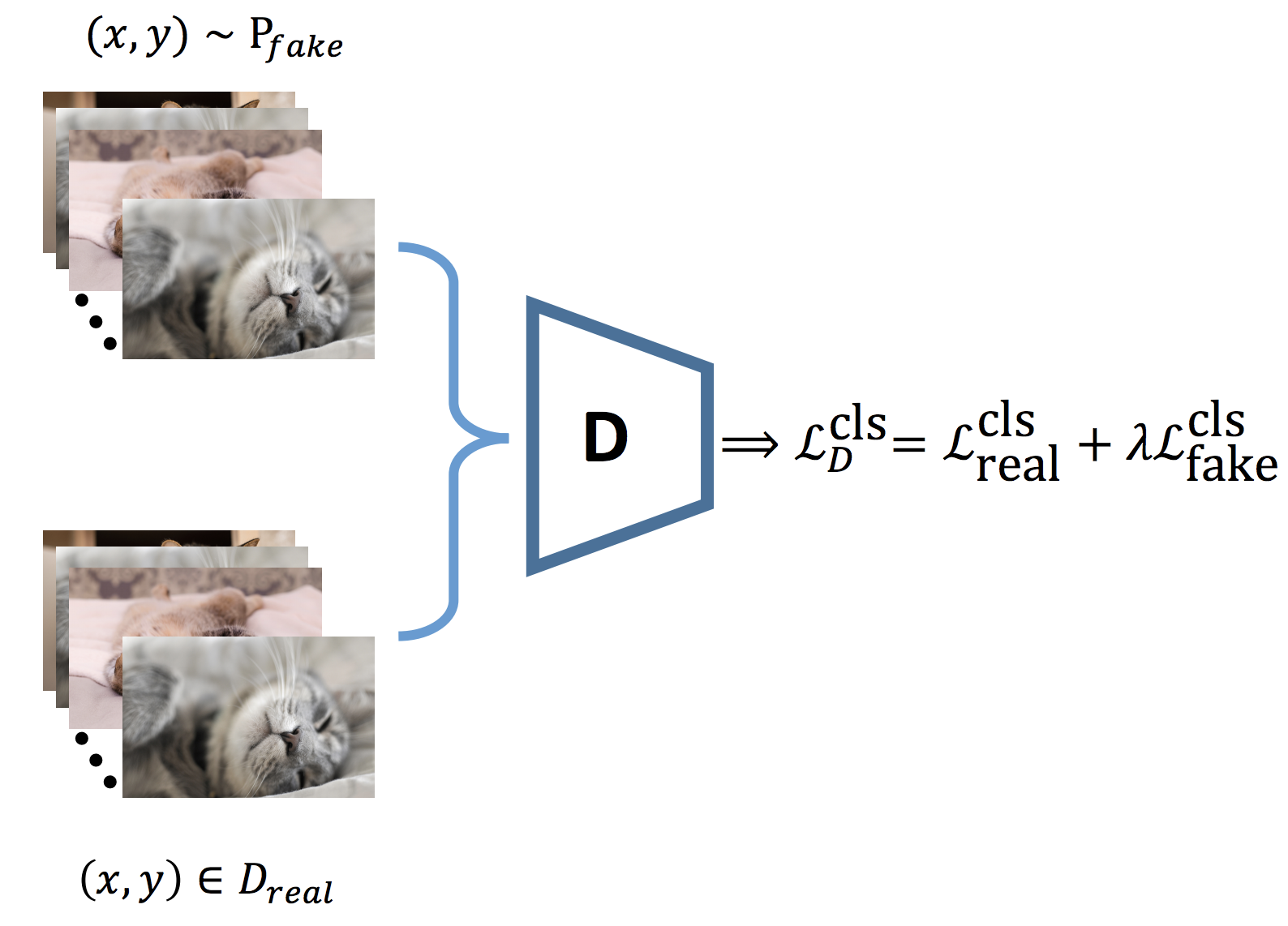}
    \vspace{-5pt}
    \caption{Illustration of fine-tuning the discriminator. We omit the adversarial attack here for brevity.}
    \label{fig:fine-tune-diagram}
    \vspace{-5pt}
\end{figure}
After end-to-end training, the discriminator has learned to minimize both \textbf{discrimination loss} and \textbf{classification loss} (see Fig.~\ref{fig:diagram}). 
If we want to train the discriminator to conduct a pure multi-class classification task, 
we will need to fine-tune it by combining fake and real images and conducting several steps of SGD only on the robustness classification loss (illustrated in Fig.~\ref{fig:fine-tune-diagram}):
\begin{equation}
\vspace{-3pt}
    \label{eq:fine-tune-loss}
    \begin{aligned}
    &\mathcal{L}_{D}^{\text{cls}}\triangleq
    \mathop{\mathbb{E}}_{(x,y)\sim\mathcal{P}_{\text{real}}}\ell(f(x_{\text{adv}};w),y)+\\&\qquad\lambda\cdot \mathop{\mathbb{E}}_{(x,y)\sim\mathcal{P}_{\text{fake}}}\ell(f(x_{\text{adv}};w),y),
    \end{aligned}
    \vspace{-3pt}
\end{equation}
where $x_{\text{adv}}=\argmin_{\|x'-x\|\le\delta_{\max}}\ell(f(x';w),y)$. Here the function $f(x; w)$ is just the classifier branch of discriminator $D(x)$, recalling that we are dealing with conditional GAN. As we can see, throughout the fine-tuning stage, we force the discriminator to focus on the classification task rather than the discrimination task. The experiments will show that the fine-tuning step improves the accuracy by a large margin. 

\section{Experimental Results}
\begin{figure*}[htb]
    \centering
    \includegraphics[width=0.3\linewidth]{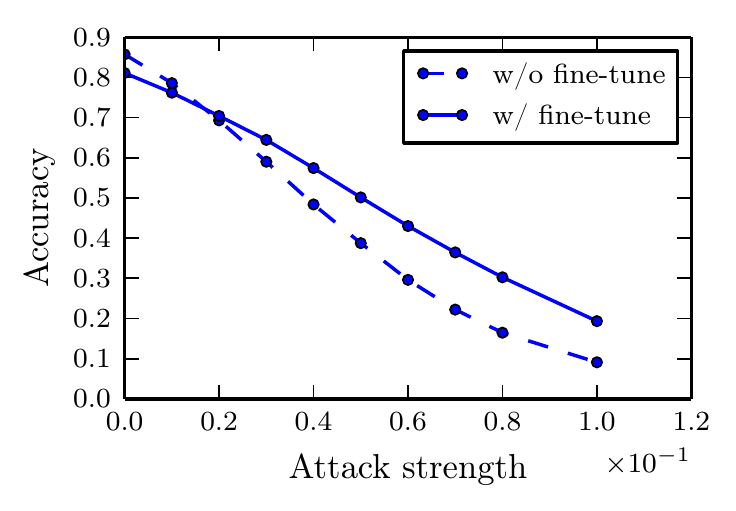}
    \includegraphics[width=0.312\linewidth]{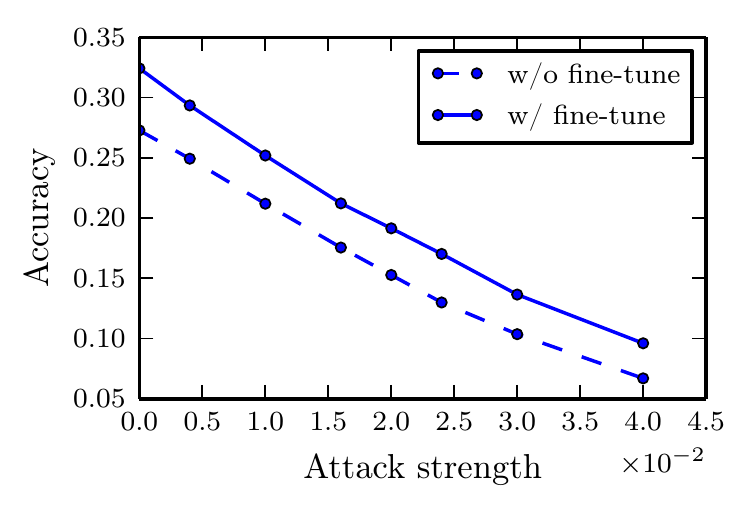}
    \includegraphics[width=0.35\linewidth]{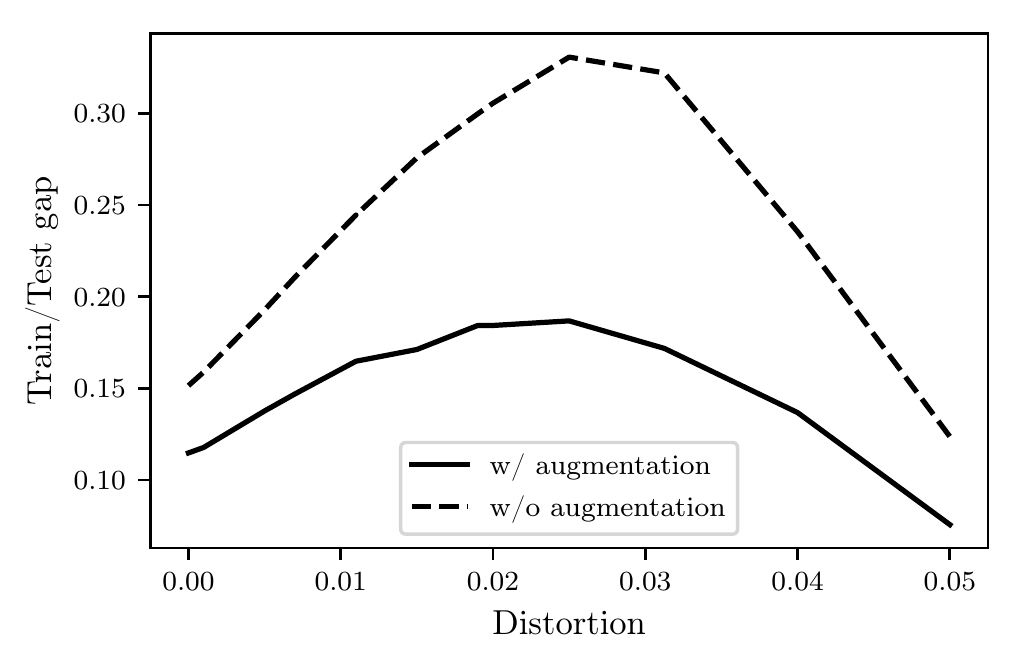}
    \vspace{-5pt}
    \caption{\textit{Left two panels}: The effect of fine-tuning on prediction accuracy (\textit{left:} CIFAR10, \textit{middle:} ImageNet-64px). \textit{Right panel}: Comparing the accuracy gap between adversarial training model and GAN data augmentation model.}
    \label{fig:fine-tune}
\vspace{-10pt}\end{figure*}
\label{sec:experiment}
We experiment on both CIFAR10 and a subset of ImageNet data. Specifically, we extract classes $y_i$ such that $y_i\in \texttt{np.arange(151, 294, 1)}$ from the original ImageNet data: recall in total there are~$1000$ classes in ImageNet data and we  sampled~$294-151=143$ classes from them. We choose these datasets because 1) the current state-of-the-art GAN,  SN-GAN~\cite{miyato2018cgans}, also worked on these datasets, and 2) the current state-of-the-art adversarial training method~\cite{madry2017towards} cannot scale to ImageNet-1k data. In order to have a fair comparison, we copy all the network architectures of generators and discriminators from SN-GAN. Other important factors, such as learning rate, optimization algorithms, and number of discriminator updates in each cycle are also kept the same. The only modification is that we discarded the feature projection layer and applied the auxiliary classifier~(see Fig.~\ref{fig:discriminators}).

\subsection{Quality of Discriminator}

We show that \texttt{Rob-GAN} leads to more robust discriminator than state-of-the-art adversarial trained models. 

\vspace{-10pt}
\paragraph{Effect of fine-tuning.}
We first compare \texttt{Rob-GAN} with/without fine-tuning
to verify our claim in Sec.~\ref{sec:augment} that fine-tuning improves classification accuracy.
To this end, we compare two sets of models: in the first set, we directly extract the auxiliary classifiers from discriminators to classify images; in the second set, we apply fine-tuning strategy to the pretrained model as Fig.~\ref{fig:fine-tune-diagram} illustrated.  The results are in Fig.~\ref{fig:fine-tune} (left), which suggests that fine-tuning is useful. 
\vspace{-10pt}
\paragraph{Accuracy gap comparison: with or without data augmentation.}
We check whether adversarial training with fake data augmentation~\eqref{eq:aug_semi_loss} really shrinks the generalization gap. To this end, we draw the same figure as Fig.~\ref{fig:accuracy-gap}, except that now the classification model is the discriminator of Rob-GAN with fine tuning. 
We compare the accuracy gap in Fig.~\ref{fig:fine-tune} (\textit{Right}). Clearly the model trained with the adversarial \textit{real+fake augmentation} strategy works extremely well: it improves the testing accuracy under PGD-attack and so the generalization gap between training/testing set does not increase that much.
\vspace{-15pt}
\paragraph{Robustness of discriminator: comparing robustness with/ without data augmentation.}
\begin{table}[htb]
\tabcolsep=0.11cm
    \centering
    \begin{threeparttable}
    \scalebox{0.8}{
    \begin{tabular}{c|c|cccc}
    \toprule
    \multirow{2}{*}{Dataset}       &
    \multirow{2}{*}{Defense}       &
    \multicolumn{4}{c}{$\delta_{\max}$ of $\ell_{\infty}$ attacks}\\\cmidrule{3-6}
                &  & $0$ & $0.02$ & $0.04$ & $0.08$ \\
    \midrule
    \multirow{2}{*}{CIFAR10} &  Adv. training  & \textbf{81.45\%} & 69.15\% & 53.74\% & 23.58\% \\
     & \texttt{Rob-GAN} (w/ FT) & 81.1\% & \textbf{70.41\%} & \textbf{57.43\%} & \textbf{30.25\%}

    \\\midrule\midrule
            &      & $0$ & $0.01$ & $0.02$ & $0.03$ \\\midrule
\multirow{2}{*}{\thead{ImageNet\textsuperscript{\textdagger}\\(64px)}}
    & Adv. Training & 20.05\% & 18.3\% & 12.52\% & 8.32\% \\
    & \texttt{Rob-GAN} (w/ FT) & \textbf{32.4\%} & \textbf{25.2\%} & \textbf{19.1\%} & \textbf{13.7\%} \\
    \bottomrule
    \end{tabular}
    }
    \begin{tablenotes}
    \item \textsuperscript{\textdagger}Denotes the 143-class subset of ImageNet.
    \end{tablenotes}
    
    \caption{\label{tab:augmentation}Accuracy of our model under $\ell_{\infty}$ PGD-attack. ``FT'' means fine-tuning.}
    \end{threeparttable}
\end{table}
\vspace{-10pt}
In this experiment, we compare the robustness of discriminators trained by Rob-GAN with the state-of-the-art adversarial training algorithm by~\cite{madry2017towards}. 
As shown in a recent comparison~\cite{athalye2018obfuscated}, adversarial training~\cite{madry2017towards} achieve state-of-the-art performance in terms of robustness under adversarial attacks. 
Since adversarial training is equivalent to Rob-GAN without the GAN component, for fair comparison we keep all the other components (network structures) the same. 

To test the robustness of different models,  we choose the widely used $\ell_{\infty}$ PGD attack~\cite{madry2017towards}, but other gradient based attacks are expected to yield the same results. We set the $\ell_{\infty}$ perturbation to $\delta_{\max}\in \texttt{np.arange(0, 0.1, 0.01)}$ as defined in \eqref{eq:PGD-attack}. Another minor detail is that we scale the images to $[-1, 1]$ rather than usual $[0, 1]$. This is because generators always have a $\tanh()$ output layer, so we need to do some adaptations accordingly. We present the results in Tab.~\ref{tab:augmentation}, which clearly shows that our method (\texttt{Rob-GAN} w/ FT) performs better than state-of-the-art defense algorithm.

\subsection{Quality of Generator}
\begin{figure*}[h]
    \centering
    \vspace{-10pt}
    \includegraphics[width=0.35\linewidth]{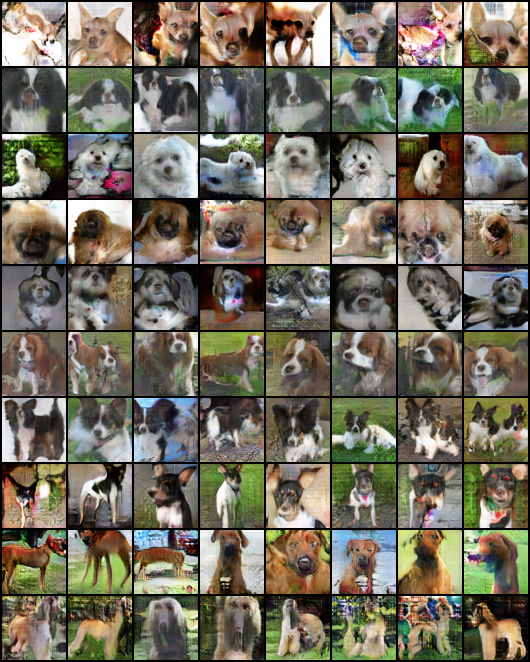}
    \includegraphics[width=0.35\linewidth]{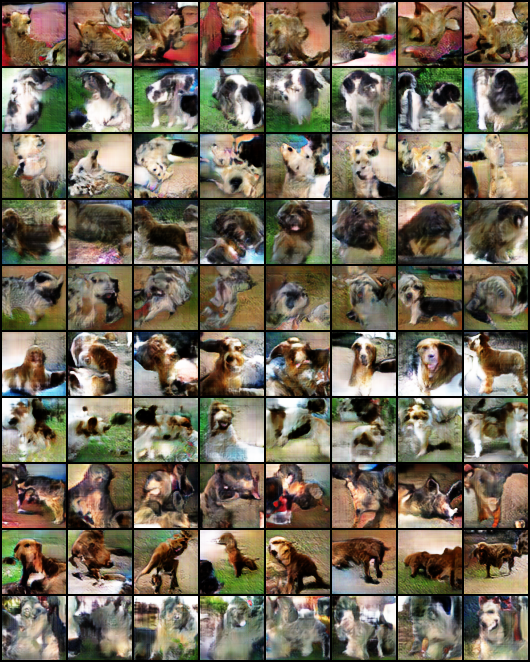}
    \caption{Comparing the generated images trained by our modified loss(\textit{left}) with the original AC-GAN loss(\textit{right}). For fair comparison, both networks are trained by inserting adversarial attacker (Sec.~\ref{sec:gan-training}). We can see images from AC-GAN loss are more distorted and harder to distinguish.}
    \label{fig:loss_compare_samples}
    \vspace{-10pt}
\end{figure*}

Next we show that by introducing adversarial attack in GAN training, \texttt{Rob-GAN} improves the convergence of the generator. 
\vspace{-10pt}
\paragraph{Effect of split classification loss. }
Here we show the effect of split classification loss described in \eqref{eq:L_C_splitted}. Recall that if we apply the loss in \eqref{eq:loss_acgan_origin} then the resulting model is AC-GAN. It is known that AC-GAN can easily lose modes in practice, i.e. the generator simply ignores the noise input $z$ and produces fixed images according to the label $y$. This defect is observed in many previous works~\cite{huang2017stacked,mathieu2015deep,isola2017image}. In this ablation experiment, we compare the generated images trained by two loss functions in Fig.~\ref{fig:loss_compare_samples}. Clearly the proposed new loss outperforms the AC-GAN loss. 
\vspace{-10pt}
\begin{figure*}[htb]
    \centering
    \includegraphics[width=0.9\linewidth]{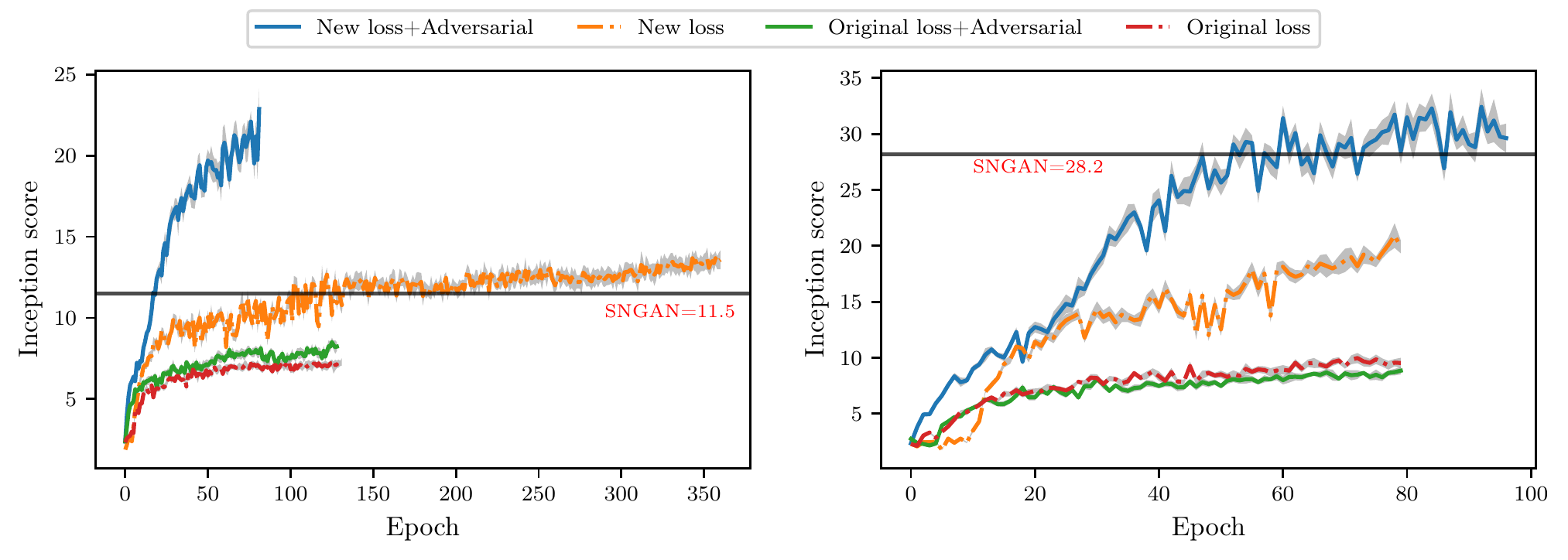}
    \vspace{-10pt}
    \captionof{figure}{Results on subset of ImageNet, left: 64px, right: 128px. Here we tried four combinations in total: with or without adversarial training, new loss or original loss. We have three findings: 1) Compared with SN-GAN, our model (new loss + adversarial) learns a high quality generator efficiently: in both datasets, our model surpasses SN-GAN in just 25 epochs (64px) or 50 epochs (128px). 2) When comparing the new loss with the original loss, we see the new loss performs better. 3) Using the new loss, the adversarial training algorithm has a great acceleration effect.}
    \label{fig:inception_convergence}
    \vspace{-10pt}
\end{figure*}

\paragraph{Quality of generator and convergence speed.}
Finally, we evaluate the quality of generators trained on two datasets: ImageNet subset - 64px and ImageNet subset - 128px. We compare with the generator obtained by SN-GAN, which has been recognized as a state-of-the-art  conditional-GAN model for learning hundreds of classes. Note that SN-GAN can also learn the conditional distribution of the entire ImageNet data ($1000$ classes), unfortunately, we are not able to match this experiment due to time and hardware limit. 
To show the performance with/without adversarial training and with/without new loss, we report the performance of all the four combinations in Figure~\ref{fig:inception_convergence}. 
Note that ``original loss'' is equivalent to AC-GAN.
Based on Figure~\ref{fig:inception_convergence} we can make the following three observations. First, adversarial training can improve the convergence speed of GAN training and make it converge to a better solution. Second, the new loss leads to better solutions on both datasets. Finally, the proposed Rob-GAN outperforms SN-GAN (in terms of inception score) on these two datasets.

\section{Conclusions}
We show the generator can improve adversarial training, and the adversarial attacker can improve GAN training. Based on these two insights, we proposed to combine generator, discriminator and adversarial attacker in the same system and conduct end-to-end training. The proposed system simultaneously leads to a better generator and a more robust discriminator compared with state-of-the-art models. 
\section*{Acknowledgments}
We acknowledge the support by NSF IIS1719097, Intel, Google Cloud and AITRICS.

{\small
\bibliographystyle{ieee}
\bibliography{ref}
}

\end{document}

%% file: math_commands.tex
\usepackage{amsmath,amsfonts,bm}

\def\1{\bm{1}}

\DeclareMathAlphabet{\mathsfit}{\encodingdefault}{\sfdefault}{m}{sl}
\SetMathAlphabet{\mathsfit}{bold}{\encodingdefault}{\sfdefault}{bx}{n}

\DeclareMathOperator*{\argmax}{arg\,max}
\DeclareMathOperator*{\argmin}{arg\,min}